\documentclass{article}

\usepackage[preprint]{neurips_2020}

\usepackage[utf8]{inputenc} 
\usepackage[T1]{fontenc}    
\usepackage{hyperref}       
\usepackage{url}            
\usepackage{booktabs}       
\usepackage{amsfonts}       
\usepackage{nicefrac}       
\usepackage{microtype}      
\usepackage{natbib}

\usepackage{graphicx}
\usepackage{amsfonts}
\usepackage{float}
\usepackage{amsmath}	
\usepackage{subcaption}
\usepackage{enumerate}
\usepackage{verbatim}
\usepackage{soul}
\usepackage{mathtools}
\usepackage{amssymb}
\usepackage{bm}
\usepackage{enumitem}
\usepackage[linesnumbered, ruled,vlined]{algorithm2e}
\usepackage{mleftright, xparse}
\usepackage{tabularx}

\newcommand{\Espace}{\mathcal{E}}

\newcommand{\Xspace}{\mathcal{X}}
\newcommand{\vecb}{\vec{b}}
\newcommand{\vecd}{\vec{d}}
\newcommand{\hatd}{\hat{d}}
\newcommand{\vecf}{\vec{f}}
\newcommand{\hatf}{\hat{f}}
\newcommand{\vectau}{\vec{\tau}}
\newcommand{\vecchi}{\vec{\chi}}
\newcommand{\hatchi}{\hat{\chi}}

\newcommand{\Map}{\mathcal{M}}

\newcommand{\argmin}{\operatornamewithlimits{argmin}}

\newcommand{\hak}[1]{\left[ #1 \right]}

\newcommand{\tes}[1]{\left( #1 \right)}
\newcommand{\braces}[1]{\left\{ #1 \right\}}

\newcommand{\beq}{\begin{equation}}
\newcommand{\eeq}{\end{equation}}
\newcommand{\beqa}{\begin{eqnarray}}
\newcommand{\eeqa}{\end{eqnarray}}

\NewDocumentCommand{\evalat}{sO{\big}mm}{%
  \IfBooleanTF{#1}
   {\mleft. #3 \mright|_{#4}}
   {#3#2|_{#4}}%
}

\title{Supervised learning on heterogeneous, attributed entities interacting over time}

\author{
  Amine Laghaout \\
  CSIS Security Group A/S, \\ 
  Vestergade 2B, 4. sal, \\
  Copenhagen K, Denmark
}

\begin{document}

\maketitle

\begin{abstract}
Most physical or social phenomena can be represented by ontologies where the constituent entities are interacting in various ways with each other and with their environment. Furthermore, those entities are likely heterogeneous and attributed with features that evolve dynamically in time as a response to their successive interactions. In order to apply machine learning on such entities, e.g., for classification purposes, one therefore needs to integrate the interactions into the feature engineering in a systematic way. This proposal shows how, to this end, the current state of graph machine learning remains inadequate and needs to be be augmented with a comprehensive feature engineering paradigm in space and time.
\end{abstract}

\section{Introduction}
\label{sec:introduction}

\subsection{Motivation}
\label{sec:Motivation}

In the industry, and even in the scientific literature, supervised learning has overwhelmingly been applied to data sets where the training examples are
\begin{enumerate}[label=({\roman*})]
\item independent of each other, and \label{pt:independent}
\item homogeneous, i.e., the subjects of the classfication or regression are instances of the same entity type such that each column in the design matrix has a consistent interpretation and format across all rows. \label{pt:homogeneous}
\end{enumerate}

In other words, the training examples used for the estimation\footnote{Estimation shall herein refer indiscriminately to either classification or regression.} refer to entities that are \ref{pt:independent} non-interacting and of \ref{pt:homogeneous} the same type. This assumption---or rather, approximation---implies that the entities are decoupled from their environment and that their \href{https://en.wikipedia.org/wiki/Design_matrix}{design matrix} is self-contained. In the real world, however, one cannot make such an assumption since any given entity to be estimated is likely part of a broader \href{https://en.wikipedia.org/wiki/Ontology_(information_science)}{ontology} which intertwines its properties with those of other entities. 

One can attempt to handcraft the relationships of the ontology into the design matrix of each entity type, but such a feature engineering exercise is demanding in human expertise, prone to the introduction of biases, and more importantly, cannot generalize to arbitrary problem domains. Significant progress in overcoming this limitation has been afforded by the recent rise to prominence of graph machine learning (GML) \citep{wu2020comprehensive, zhang2018deep}. Applications of GML to evidently graph-based ontologies such as social networks \citep{al-eidi2020time-ordered} or cybersecurity \citep{liu2019heterogeneous, sun2019hindom} have quickly flourished and developer tools such as  \href{https://www.stellargraph.io/}{StellarGraph} \citep{StellarGraph} are reaching maturity. Despite all these advances, a thoroughly comprehensive and generalizable approach has yet to be devised for the estimation of heterogeneous, attributed nodes that interact in time. This latter scenario is indeed the most ubiquitous in the real world since, at the most fundamental level, all phenomena can be reduced to physical interactions between indivisible entities. Some attempts in this direction, such as spatio-temporal graphs \citep{zhang2018gaan}, do take into account the space-time aspect of the problem but cannot accommodate the heterogeneity of the graph. Conversely, the few algorithms that can handle heterogeneity such as GraphSAGE \citep{hamilton2017inductive}, and its derivative HinSAGE, cannot model time-evolution of both the node attributes and of the edges. More crucially, the main shortcoming of GML lies in the fact that interactions between entities are constrained to bi-partite relationships. This makes it inapplicable for problems where the interaction can---in principle---behave as a stochastic black box involving an arbitrary number of vertices with no particular directionality. 

This proposal aims to resolve the above problem by outlining a reference architecture for the estimation of heterogeneous, attributed entities that interact with their environment---and with each other---over time. In order to accommodate arbitrary interactions, it shall do away with GML's attempts to force-fit\footnote{One particular workaround that could use GML techniques shall be ignored here, namely the one where interactions are not edges, but vertices, on an equal footing with entities. Notwithstanding the fact that blending interactions and entities in the same set of vertices is conceptually inelegant, the proliferation of nodes that would ensue would lead to unwieldy dimensionalities.} graph topologies onto the ontology of the problem domain. Instead, it shall let interactions be modeled as learnable modules that can be recycled for any entity instances they involve. Because a parallel can be drawn between interactions and the notion of \href{https://en.wikipedia.org/wiki/Hypergraph}{hyperedges}, one can describe the proposal herein as a revised, tentative blueprint for \textit{hypergraph machine learning} \citep{zhou2006learning, jiang2019dynamic}.

The building blocks of the problem are formally defined in \S\ref{sec:Building blocks}; \S\ref{sec:Supervised learning} presents the architecture and spells out the learning problem in terms of the building blocks; and \S\ref{sec:Outlook} goes over the open questions and blind spots that may require further investigation.

\subsection{A real-world use case: COVID-19}
\label{sec:A real-world use case: COVID-19}

An intuition for the power and ubiquity of the present proposal is best elicited by a real-world use case. Due to its global impact and media exposure, a relatable application is the modeling of the spread of a pandemic, such as COVID-19. The goal of the model is to determine whether an \textit{entity} is infected by---or acts as a vector for---the disease. Here, \textit{entities} could be biological (e.g., human beings, pets, wild animals) or inanimate (e.g., objects such as dooknobs or handrails, or venues such as markets or fitness clubs). One can see that each of these entities have attributes, i.e., features, that are intrinsic to them. These can be the age or genetic makeup for a biological entity, or the capacity or level of sanitation for a physical venue. Note how these \textit{intrinsic features} are most often dynamic and hence require a temporal treatment. \textit{Extrinsic features}, on the other hand, arise from the \textit{interactions} that the entities have had with one another, conditioned on various \textit{environmental parameters}. As entities interact, their extrinsic features---which are also dynamic---are to be updated so as to reflect the changing likelihood that any given entity carries the virus.\footnote{Note how the plethora of preventative measures that were taken during the pandemic, such as social distancing, lockdowns, or the adoption of face masks, are all modulations on the environemental parameters, or even direct changes to the intrinsic and extrinsic features, aimed at minimizing the likelihood that entities are classified as vectors of the disease.}

\section{Building blocks}
\label{sec:Building blocks}

\subsection{Entities}
\label{sec:Entities}

Let $\varepsilon_{k}^{(j)}$ be the $k$-th instance of an entity of type $j$. The set $\Espace^{(j)}$ of entities of type $j$ adds up to the overall, hetergeneous set $\Espace$ of all entities, i.e.,
\beq
\varepsilon_{k}^{(j)} \in \Espace^{(j)} \subset \Espace = \bigcup\limits_{l=1}^{E} \Espace^{(l)},
\eeq
where $E$ is the number of entity types. Each entity $\varepsilon_{k}^{(j)}$ is represented algebraically by a vector of features $\hatd_k^{(j)}$ whose interpretation and dimensionality is fixed by the entity type (cf. \S\ref{sec:Data representation}). 

\subsection{Interactions}
\label{sec:Interactions}

Entities are not static in the sense that they interact with their environment, thereby updating their features upon those interactions.  The most trivial interaction involves a single entity $\varepsilon$ whose labels or attribute features are re-written by the environment independently of other entities. A more interesting case happens when the environment also involves one or more other entities $\varepsilon' \neq \varepsilon$ of potentially different types. These other entities $\varepsilon'$ both influence---and are in turn influenced by---the presence of $\varepsilon$. Such interactions thus induce correlations (i.e., dependencies) or perturbations (i.e., noise) among the features of the entities at play.

Let us formalize the above by denoting the $l$-th instance of an interaction of type $i$ as a function
\beq
\chi_{l}^{(i)} = \chi_{l}^{(i)}(\xi_l^{(i)}, \vectau_l^{(i)}, t)
\label{eq:interaction}
\eeq
of the set\footnote{$\xi_l^{(j)}$ is, strictly speaking, a dictionary, or associative array, of key-value pairs where the key specifies the role in the interaction, and the value specifies the entity instance.} $\xi_l^{(i)} \subset \Espace$ of entities involved, the vector $\vectau_l^{(i)}$ of environmental parameters that modulate the interaction, and the timestamp $t$ of when the interaction occurred. Note that a given interaction type $i$ predetermines the structure of both $\xi_l^{(i)}$ and $\vectau_l^{(i)}$. One can think of an interaction \textit{type} as a set of co-occurring relationships within the broader ontology of the problem domain. An interaction is instantiated, i.e., subscripted with some index $l$ as in Eq. (\ref{eq:interaction}), once it is time-stamped and associated with a particular set of entity instances $\xi_l^{(j)}$ and environmental parameters $\vectau_l^{(i)}$.

Just as for entities, interactions can be grouped into sets according to their types, thereby adding up to the overall set $\Xspace$ of $I$ possible interaction types:
\beq
\chi_{l}^{(i)} \in \Xspace^{(i)} \subset \Xspace = \bigcup\limits_{\iota=1}^{I} \Xspace^{(\iota)}.
\label{eq:interaction sets}
\eeq

Notice how $\chi$ is the multipartite generalization of the attributed bipartite edge $\xi = \braces{\varepsilon, \varepsilon'}$ commonly known from ``run-of-the-mill'' graph theory. The definition of an interaction in Eq. (\ref{eq:interaction}) is thus  more akin to a hyperedge that spans $\xi$, is attributed with $\vectau$, and is time-stamped at $t$.

\subsection{Data representation}
\label{sec:Data representation}

Let the knowledge about an entity $\varepsilon_k^{(j)}$ at time $t$ be encoded by its \textit{data vector}\footnote{or more generally, a tensor} 
\beqa
\hatd_k^{(j)}(t) & = & \overbrace{\vecb_k^{(j)}(t)}^{\scriptsize \mbox{targets}} \oplus \overbrace{\hatf_k^{(j)}(t)}^{\scriptsize \mbox{intrinsic features}} \oplus \hak{\overbrace{\bigoplus\limits_{\forall i\mid\varepsilon^{(j)}\in \xi^{(i)}}  \hatchi_k^{(j, i)}(t)}^{\scriptsize\mbox{extrinsic features}}},
\label{eq:data vector}
\eeqa
which is the concatenation\footnote{Concatenation shall be symbolized mathematically as the direct sum operator $\oplus$.} of three vectors, namely the target features, the intrinsic features, and the extrinsic features. Notice that the extrinsic features are themselves a concatenation of as many interactions as an entity of type $j$ is involved in. This is developed further in \S\ref{sec:Extrinsic features}.

\subsubsection{Target features}
\label{sec:Target features}

The \textit{target features} $\vecb_k^{(j)}(t)$ at time $t$ of an entity $\varepsilon_k^{(j)}$ are the scores or labels\footnote{One  shall refer to scores and labels interchangeably, thus leaving the freedom to the particular use-case to decide whether it is a matter of regression or classification, respectively.} which drive supervised learning. One can assume that one begins with a body of labeled entities for which the targets are well-defined and serve as ground truths or, more realistically, as seed \textit{beliefs}\footnote{hence the $b$-notation for beliefs} for the estimation of the entities.

\subsubsection{Intrinsic features}
\label{sec:Intrinsic features}

The \textit{intrinsic features} $\hatf_k^{(j)}(t)$ at time $t$ of an entity $\varepsilon_k^{(j)}$ are any features which can be completely decoupled from the presence of other entities $\varepsilon_{k'}^{(j')}\neq\varepsilon_k^{(j)}$ in the environment. Moreover, since the system is dynamic, $\hatf_k^{(j)}$ is an aggregation through time of all the sequential updates $\vecf_k^{(j)}(t'\mid t' < t)$ undergone by $\varepsilon_k^{(j)}$ up to time $t$. While $\vecf_k^{(j)}$ is most often human-readable, $\hatf_k^{(j)}$ can instead be an abstract encoding that collapses the history of intrinsic feature updates onto a fixed, lower-dimensional space. This time-collapse---or aggregation---operation $\Map_f^{(j)}$ shall be denoted by 
\beqa
\hatf_k^{(j)}(t) & = & \Map_f^{(j)}\!\!\tes{\bigoplus\limits_{t-\Delta T \leq t' < t}^{t} \hak{\vecf_k^{(j)}(t'), t'}}
\label{eq:intrinsic-time-aggregator-expanded}
\eeqa
where $\Delta T$ is the lookback period from the current time $t$ and should ideally span all the way back to the creation time of $\varepsilon_k^{(j)}$.\footnote{Because of implementational constraints, however, only the most recent interval of history can be stored in memory so $\Delta T$ will most likely be a finite time window.} Note that the \textit{time-aggregator} $\Map_f^{(j)}$ of intrinsic features does not merely operate on the unordered set of updates $\vecf_k^{(j)}(t')$, but rather on their \textit{history}, i.e., on pairs $\hak{\vecf_k^{(j)}(t'), t'}$ where $t'$ is needed to serve as an \textit{attention} parameter.\footnote{More recent events typically deserve more attention than older ones.} One can thus re-write Eq. (\ref{eq:intrinsic-time-aggregator-expanded}) as 
\beq
\hatf_k^{(j)}(t) = \Map_f^{(j)}\tes{H(\vecf_k^{(j)})\Big|_{\scriptsize t{-}\Delta T}^{\scriptsize t}}
\label{eq:intrinsic-time-aggregator}
\eeq
where 
\beq
H(v)\Big|_{\scriptsize t{-}\Delta T}^{\scriptsize t} = \bigoplus\limits_{t-\Delta T \leq t' < t}^{t} \hak{v(t'), t'}
\label{eq:history}
\eeq
represents the time-stamped history of any variable $v$ from $t{-}\Delta T$ to $t$.

In terms of deep learning architecture, $\Map_f^{(j)}$ can be implemented by a recurrent neural network or a transformer \citep{vaswani2017attention}. However, a (psuedo-)Markovian simplification,\footnote{The viability of this simplification is subject to experimentation.} denoted $\tilde{\Map}_f^{(j)}$, could make use of only the current update and the previous aggregation, i.e., 
\beq
\hatf_k^{(j)}(t) = \tilde{\Map}_f^{(j)}\!\!\tes{\vecf_k^{(j)}(t), \hatf_k^{(j)}(t{-}1)}.
\label{eq:intrinsic-time-aggregator-Markovian}
\eeq

\subsubsection{Extrinsic features}
\label{sec:Extrinsic features}

The \textit{extrinsic features} $\hatchi_k^{(j,i)}(t)$ at time $t$ of an entity $\varepsilon_k^{(j)}$ are those that depend on the data vectors of other entities in the context of an interaction $\chi_l^{(i1)}$ of type $i$. $\hatchi_k^{(j,i)}(t)$ is a thus latent representation which summarizes the sequence of interactions of type $i$ which $\varepsilon_k^{(j)}$ has been involved in up to and including time step $t$. In a manner similar to Eqs. (\ref{eq:intrinsic-time-aggregator-expanded}, \ref{eq:intrinsic-time-aggregator}, \ref{eq:history}), this time-aggregation can be expressed by
\beq
\hatchi_k^{(j, i)}(t) = \Map_f^{(j)}\tes{H(\vecchi_k^{(j, i)})\Big|_{\scriptsize t{-}\Delta T}^{\scriptsize t}}
\label{eq:extrinsic-time-aggregator}
\eeq
where $\vecchi_k^{(j, i)}(t')$ is the latent representation---from the perspective of $\varepsilon_k^{(j)}$---of some particular interaction instance
\beq
\chi_{l}^{(i)}\!\tes{\xi_{l}^{(i)}, \vectau_{l}^{(i)}, t'} \mid \varepsilon_k^{(j)} \in \xi_{l}^{(i)}
\label{eq:space-aggregator}
\eeq
which took place at time $t'$. 

\begin{figure}
\centering
\includegraphics[width=.65\columnwidth]{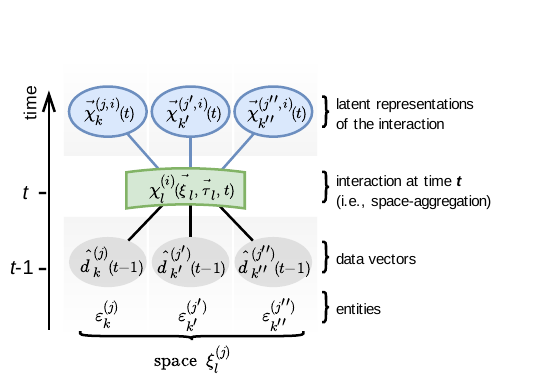}
\caption{Three entities $\xi_l^{(i)} = \{\varepsilon_{k}^{(j)}, \varepsilon_{k'}^{(j')}, \varepsilon_{k''}^{(j'')}\}$ are involved in an interaction $\chi_l^{(i)}$ of type $i$ at the time step $t$. For each of the entities, the interaction $\chi_l^{(i)}$ is represented by a vector $\vecchi$ which summarizes a ``personalized takeaway message'' from their encounter. This operation is based on the data vectors at the previous time step of the entities involved as well as a vector $\vectau_l^{(i)}$ of environmental parameters.}
\label{fig:space_aggregator}
\end{figure}

The process by which an interaction $\chi_{l}^{(i)}$ generates a latent representation $\vecchi_k^{(j, i)}$ of itself to each of its participating entities $\varepsilon_k^{(j)}$ is examplified in Fig. \ref{fig:space_aggregator} for the tri-partite case. This process shall be referred to as \textit{space-aggregation} in the sense that, unlike $\Map_f^{(j)}$ and $\Map_{\chi}^{(j, i)}$, which aggregate histories through time, $\chi_{l}^{(i)}$ aggregates the features of neighbouring entities as per the topology of the hypergraph that links them (in space). One can therefore consider $\chi_{l}^{(i)}$ as a black box for any conceivable technique from graph machine learning \citep{wu2020comprehensive} or even traditional belief propagation \citep{yedida2003understanding}. Formally, space-aggregation at time step $t$ shall be expressed as the mapping
\beqa
\chi_{l}^{(i)}: \bigcup\limits_{\varepsilon_k^{(j)} \in \xi_l^{(i)}} \braces{\vecd_k^{(j)}(t{-}1)} \rightarrow \bigcup\limits_{\varepsilon_k^{(j)} \in \xi_l^{(i)}} \braces{\vecchi_k^{(j,i)}(t)}.
\label{eq:space-aggregator-explicit}
\eeqa

Finally, going back to the time-aggregation $\Map_\chi^{(j, i)}$ of extrinsic features, one can consider the Markovian assumption
\beq
\hatchi_k^{(j, i)}(t) = \tilde{\Map}_{\chi}^{(j, i)}\!\!\tes{\vecchi_k^{(j, i)}(t), \hatchi_k^{(j, i)}(t{-}1)}.
\label{eq:extrinsic-time-aggregator-Markovian}
\eeq

\section{Supervised learning}
\label{sec:Supervised learning}

\subsection{Circuit diagram of the estimation}

\begin{figure*}
\includegraphics[width=1\columnwidth]{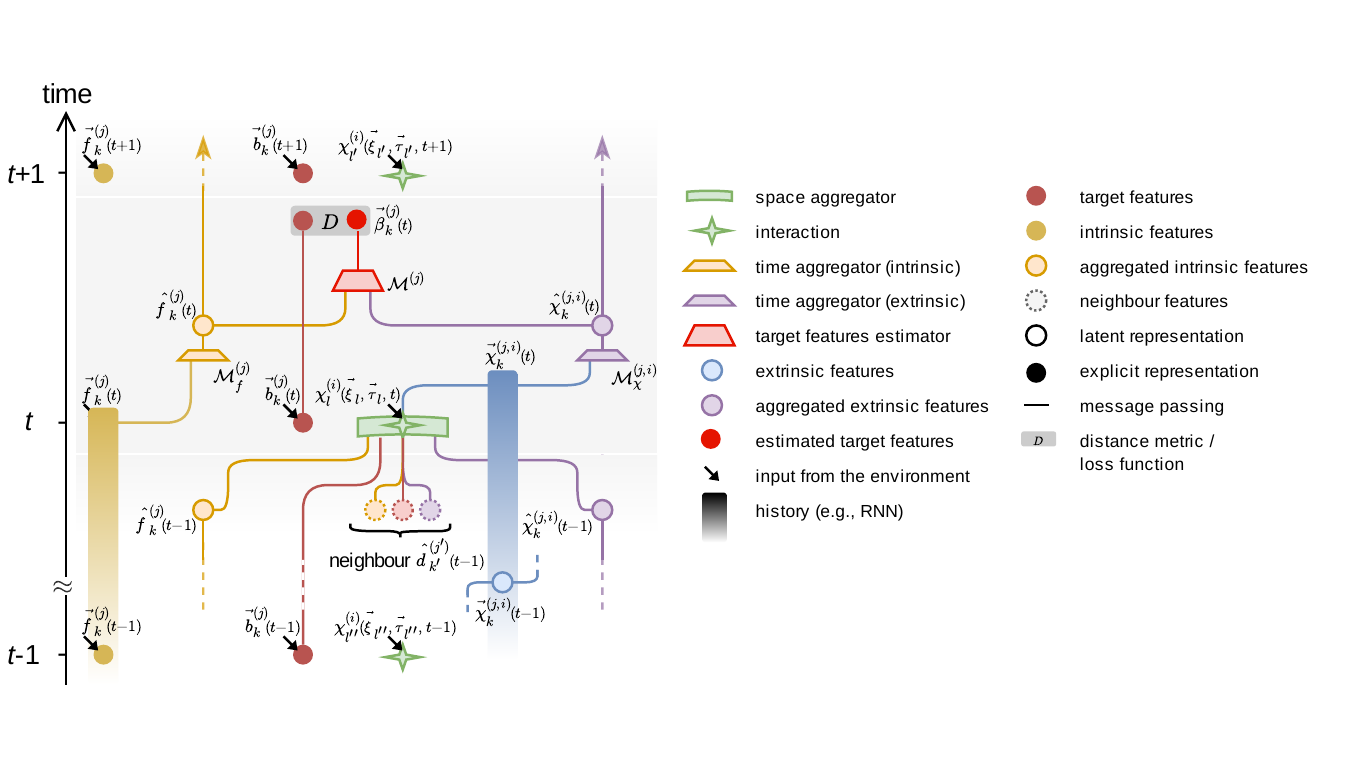}
\caption{This circuit diagram shows the successive transformations that are undergone by entity $\varepsilon_k^{(j)}$ as it interacts with its environment on a discretized time scale. Three possible events---i.e., inputs from the environment---can occur at any time step $t$, namely the update of its target features, the update of its intrinsic features, or its involvement in an interaction with neighbouring entities. For each of these events, the data vector of $\varepsilon_k^{(j)}$ is re-processed by direct overwriting (of the targets), by time-aggregation $\Map_{f}^{(j)}$ (of the intrinsic features), or by space-aggregation $\chi_{l}^{(j)}$ followed by time-aggregation $\Map_{\chi}^{(j, i)}$ (of the extrinsic features). The resulting latent representation is then merged by an overarching mapping $\Map^{(j)}$ which projects it on the same space as that of the target features. All four mappings $\Map_{f}^{(j)}$, $\chi_{l}^{(i)}$, $\Map_{\chi}^{(j, i)}$, and $\Map^{(j)}$ are therefore to be optimized in view of a single common goal, namely the minimization Eq. (\ref{eq:optimization}) of the loss function $D$. For simplicity, only the swimlane relevant to $\varepsilon_k^{(j)}$ is shown here and all irrelevant connections onto the swimlanes of other entities are omitted (e.g., connections to $\vecchi_{k'}^{(j', i)}$). Similarly, in order to reduce clutter, only the input at time $t{-}1$ from a single neightbour $\varepsilon_{k'}^{(j')}$ is shown, although, in practice, any number of entities can converge at the interaction node $\chi_l^{(j)}$. Finally, once again for the sake of simplicity, only one interaction instance is shown.}
\label{fig:architecture_non_markovian}
\end{figure*}

\begin{figure*}
\includegraphics[width=1\columnwidth]{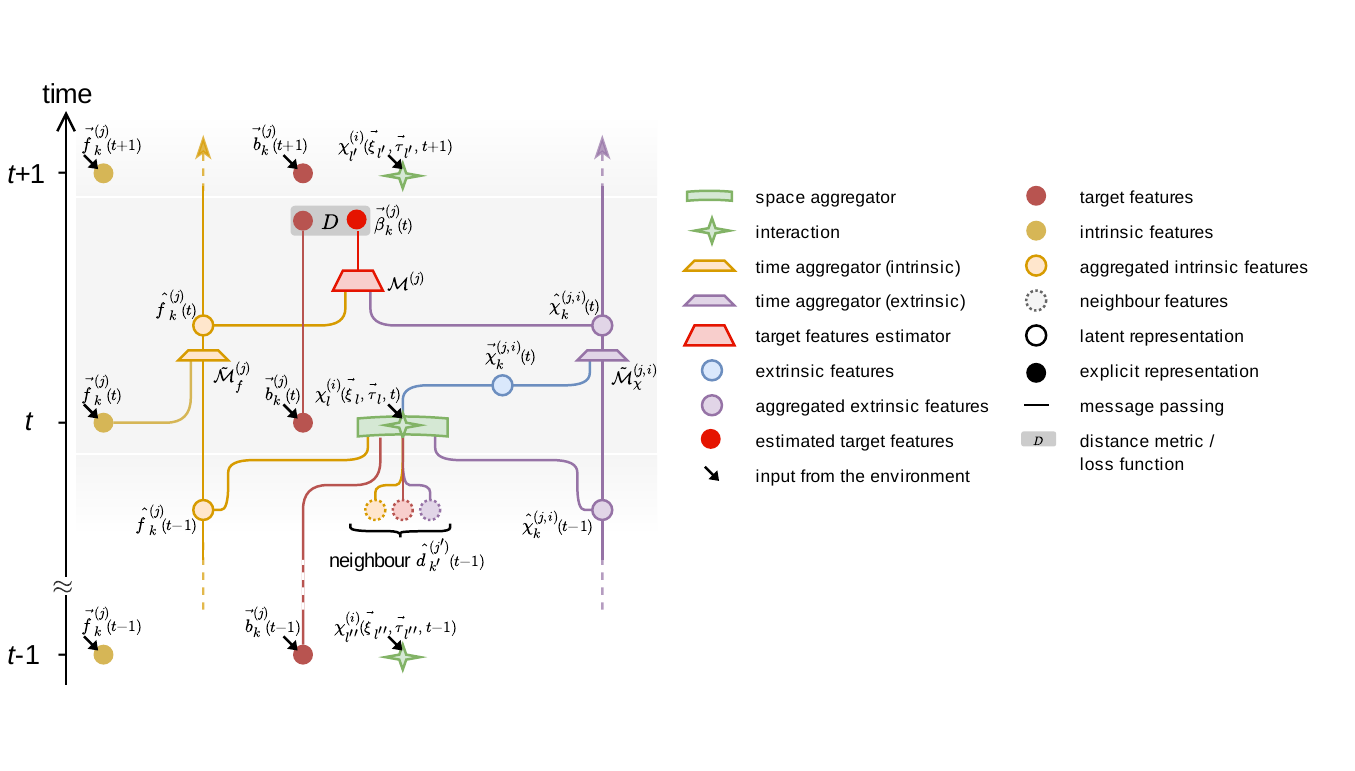}
\caption{Analog of the circuit diagram of Fig. \ref{fig:architecture_non_markovian} based on the (pseudo-)Markovian assumptions for time-aggregation, i.e., Eqs. (\ref{eq:intrinsic-time-aggregator-Markovian}) and (\ref{eq:extrinsic-time-aggregator-Markovian}).}
\label{fig:architecture_markovian}
\end{figure*}

Figure \ref{fig:architecture_non_markovian} brings together the building blocks that were presented above. It depicts, on a discretized timeline, the flow of information that culminates at time $t$ with the training on---or the serving of---a target belief for entity $\varepsilon_k^{(j)}$. Figure \ref{fig:architecture_markovian} shows the same scenario as Fig. \ref{fig:architecture_non_markovian} under the Markovian assumptions of Eqs. (\ref{eq:intrinsic-time-aggregator-Markovian}) and (\ref{eq:extrinsic-time-aggregator-Markovian}).

As indicated by the small diagonal arrows, the environment can at any time step do any one of three operations on the entity, namely
\begin{itemize}
\item update its target belief,
\item update its intrinsic features, or
\item involve it in one or more interactions, thereby updating its extrinsic features.
\end{itemize}

The pseudocode for systematically processing three potential updates in an online fashion is shown in Alg. \ref{alg:pseudocode}.

\begin{algorithm}
\SetAlgoLined
\KwResult{The parameters of $\mathcal{M}^{(j)}$, $\mathcal{M}_{f}^{(j)}$, $\mathcal{M}_{\chi}^{(j, i)}$, and $\chi_l^{(i)}$ are optimized so as to minimize the loss function $D\!\tes{\vecb^{(j)}_k(t), \vec{\beta}_k^{(j)}(t)}$ as per Eq. (\ref{eq:optimization}).}
initialize the weights of $\mathcal{M}^{(j)}$, $\mathcal{M}_{f}^{(j)}$, $\mathcal{M}_{\chi}^{(j, i)}$, and $\chi_l^{(i)}$ randomly (or via transfer learning, if applicable)\;
\ForEach{entity instance $k$ of the fixed type $j$}
{
	\ForEach{time step $t' \le t$}
	{
		\vspace{10pt}
		\tcp{Update the intrinsic features.}		
		\uIf{the intrinsic feature $\vecf_{k}^{(j)}$ is updated at time $t'$}
		{
			append the pair $\hak{\vecf_k^{(j)}(t'), t'}$ to the history $\evalat{H(\vecf_{k}^{(j)})}{t'-\Delta T}^{t'-1}$  of intrinsic updates\;
		}
		time-aggregate the history of intrinsic updates into $\hatf_k^{(j)}(t')$ with Eq. (\ref{eq:intrinsic-time-aggregator})\;

		\vspace{10pt}
		\tcp{Update the target beliefs.}
		\uIf{the belief is updated at time $t'$}
		{
			assign the new belief to $b_k^{(j)}(t')$\;
		}
		\uElse
		{
			assume that the previous belief remained unchanged, i.e., $b_k^{(j)}(t') \leftarrow b_k^{(j)}(t'{-}1)$\;		
		}
		
		\vspace{10pt}
		\tcp{Update the extrinsic features.}
		\ForEach{interaction $\chi_l^{(i)}$ of type $i$ that $\varepsilon_k^{(j)}$ can be involved in}
		{
			\uIf{$\varepsilon_k^{(j)}$ is indeed involved in $\chi_l^{(i)}$ at time $t'$}
			{
				space-aggregate the data vectors at time $t'{-}1$ of all entities involved in $\chi_l^{(i)}$ into $\vecchi_k^{(j, i)}$ with Eq. (\ref{eq:space-aggregator-explicit})\;
				append the pair $\hak{\vecchi_k^{(j)}(t'), t'}$ to the history $\evalat{H(\vecchi_{k}^{(j)})}{t'-\Delta T}^{t'-1}$  of extrinsic updates\;
			}		
			time-aggregate the history of extrinsic updates into $\hatchi_k^{(j, i)}(t')$ with Eq. (\ref{eq:extrinsic-time-aggregator})\;
		}
		\vspace{10pt}
		\tcp{Evaluate against the target.}
		concatenate the aggregated intrinsic and extrinsic features via Eq. (\ref{eq:concatenate intrinsic and extrinsic})\;
		project the resulting vector in the space of beliefs via Eq. (\ref{eq:optimization})\;
		perform back-propagation on the aggregators so as to align the projected vector with the target belief via Eq. (\ref{eq:estimation})\;
	}
}
\caption{Online training of the estimator for an entity $\varepsilon_k^{(j)}$ of type $j$ at time $t$}
\label{alg:pseudocode}
\end{algorithm}

\subsection{Analytical derivation of the estimation}

As stated in the introduction, the aim of this article is to outline a reference architecture for supervised learning on the entities. Taking entity $\varepsilon_k^{(j)}$ as an example, the goal is to learn a function $\Map^{(j)}$ which maps its features
\beq
\hatf_k^{(j)}(t) \oplus \hak{\bigoplus\limits_{\forall i \mid \varepsilon^{(j)}\in\xi^{(i)}} \hatchi_k^{(j,i)}}
\label{eq:concatenate intrinsic and extrinsic}
\eeq
onto its target $\vecb_k^{(j)}(t)$. In practice, $\Map^{(j)}$ can only achieve an approximation 
\beq
\beta_k^{(j)}(t) = \Map^{(j)}\!\!\tes{\hatf_k^{(j)}(t) \oplus \hak{\bigoplus\limits_{\forall i \mid \varepsilon^{(j)}\in\xi^{(i)}} \hatchi_k^{(j,i)}(t)}}
\eeq
of the actual target $\vecb_k^{(j)}(t)$ such that one is left with the optimization problem
\beq
\argmin_{\Map^{(j)}} \frac{1}{K^{(j)}\Delta T} \sum\limits_{t'=t{-}\Delta T}^{t} \sum\limits_{k=1}^{K^{(j)}} D\!\tes{\vecb^{(j)}_k(t'), \vec{\beta}_k^{(j)}(t')}
\label{eq:optimization naive}
\eeq
where $K^{(j)}$ is the number of entities of type $j$ and $D$ is an arbitrary distance metric which can double as a loss function.

A full expansion of $\Map^{(j)}$ in terms of the aggregator operations in time Eqs. (\ref{eq:intrinsic-time-aggregator},  \ref{eq:extrinsic-time-aggregator}) and space Eq. (\ref{eq:space-aggregator}) yields the estimated belief at time $t$
{\small
\beqa
\beta_k^{(j)} & = & \Map^{(j)}\!\!\tes{\hatf_k^{(j)}(t) \oplus  \hak{\bigoplus\limits_{\forall i \mid \varepsilon^{(j)}\in\xi^{(i)}} \!\!\!\! \hatchi_k^{(j,i)}(t)}} \\
& = & \braces{\mbox{time-aggregations Eqs. (\ref{eq:intrinsic-time-aggregator}) and (\ref{eq:extrinsic-time-aggregator})}} \nonumber\\
& = & \Map^{(j)}\!\!\tes{\Map_f^{(j)}\!\!\tes{H(\vecf_k^{(j)})\Big|_{\scriptsize t{-}\Delta T}^{\scriptsize t}} \oplus \hak{\bigoplus\limits_{\forall i \mid \varepsilon^{(j)}\in\xi^{(i)}} \!\!\!\!\!\!\! \Map_{\chi}^{(j, i)}\!\!\tes{H(\vecchi_k^{(j, i)})\Big|_{\scriptsize t{-}\Delta T}^{\scriptsize t}}}} \label{eq:non-Markovian just before space-aggregation}\\
& = & \braces{\mbox{space-aggregation Eq. (\ref{eq:space-aggregator})}} \nonumber\\
%
& = & \Map^{(j)}\!\!\tes{\underbrace{\Map_f^{(j)}\!\!\tes{\underbrace{H(\underbrace{~~\vecf_k^{(j)}~~}_{\mbox{\scriptsize intr. update}})\Big|_{\scriptsize t{-}\Delta T}^{\scriptsize t}}_{\mbox{\scriptsize hist. of intr. updates}}}}_{\scriptsize\mbox{(latent) intrinsic features}} \oplus \underbrace{\hak{\bigoplus\limits_{\forall i \mid \varepsilon^{(j)}\in\xi^{(i)}} \!\!\!\!\!\!\! \Map_{\chi}^{(j, i)}\!\!\tes{\underbrace{\evalat*{H\tes{\underbrace{\chi_l^{(i)}\!\!\tes{\bigcup\limits_{\varepsilon_{k'}^{(j')}\in\xi^{(i)}}\braces{\vecd_{k'}^{(j')}}}}_{\mbox{\scriptsize extrinsic update / interaction}}}}{\scriptsize t{-}\Delta T}^{\scriptsize t{-}1}}_{\mbox{\scriptsize history of extrinsic updates}}}}}_{\mbox{\scriptsize (latent) extrinsic features}}}, \nonumber\\
\eeqa
}
or, if one were to apply the Markovian assumption on Eq. (\ref{eq:non-Markovian just before space-aggregation}),
{\small
\beqa
\beta_k^{(j)} & = & \Map^{(j)}\!\!\tes{\tilde{\Map_f}^{(j)}\!\!\tes{\vecf_k^{(j)}(t), \hatf_k^{(j)}(t{-}1)} \oplus \hak{\bigoplus\limits_{\forall i \mid \varepsilon^{(j)}\in\xi^{(i)}} \tilde{\Map_{\chi}}^{(j, i)}\!\!\tes{\vecchi_k^{(j, i)}(t), \hatchi_k^{(j, i)}(t{-}1)}}} \\
& = & \braces{\mbox{space-aggregation}} \nonumber\\
& = & \Map^{(j)}\!\!\tes{\tilde{\Map_f}^{(j)}\!\!\tes{\vecf_k^{(j)}(t), \hatf_k^{(j)}(t{-}1)} \oplus \hak{\bigoplus\limits_{\forall i \mid \varepsilon^{(j)}\in\xi^{(i)}} \tilde{\Map_{\chi}}^{(j, i)}\!\!\tes{\chi_l^{(i)}\!\!\tes{\bigcup\limits_{\varepsilon_{k'}^{(j')}\in\xi^{(i)}}\braces{\vecd_{k'}^{(j')}(t'{-}1)}}, \hatchi_k^{(j, i)}(t{-}1)}}}. \nonumber\\
\label{eq:estimation}
\eeqa}

One can thus see that the optimization problem of Eq. (\ref{eq:optimization naive}) is not limited to the parameters and hyper-parameters of $\Map^{(j)}$ but also extends to those of the aggregators $\Map_f^{(j)}$, $\Map_{\chi}^{(j, i)}$, and $\chi^{(j)}$ such that the global optimum is given by

\beq
\argmin_{\Map^{(j)},~\bigcup\limits_i \Map_{\chi}^{(j, i)},~\Map_f^{(j)},~\chi^{(i)}} \frac{1}{K^{(j)}\Delta T} \sum\limits_{t'=t{-}\Delta T}^{t} \sum\limits_{k=1}^{K^{(j)}} D\!\tes{\vecb^{(j)}_k(t'), \vec{\beta}_k^{(j)}(t')}.
\label{eq:optimization}
\eeq

Notice how, unlike most of the literature on graph machine learning, the data aggregators are not dependent on any particular \textit{instances} of entities and interactions but only on their \textit{types} $j$ and $i$, respectively.

\subsection{Remarks on the aggregation processes}
\label{sec:Remarks}

Some remarks are in order regarding the design of Fig. \ref{fig:architecture_non_markovian}, especially about the aggregation processes.

First, there is no time-aggregation for the targets. This is motivated by the fact that the aggregators should focus on inferring the \textit{current} target $\vecb_k^{(j)}(t)$ instead of trying to reproduce its history. Except for feeding in the latest target $\vecb_k^{(j)}(t{-}1)$ to neighbouring entities via the space-aggregator, the targets of any given entity should not leak in its feature space to avoid the risk of overfitting.\footnote{The targets from the pervious time step of the entity can---and should---however be used by its neighbouring entities via the space-aggregation process.} Another reason is that, unlike pure time-series problems (e.g., stock prediction) where the target's history is itself the main feature, the present problem much more heavily entangles the entities to their environment such that, at any time step $t$, a target can be abruptly overwritten, in complete disregard for any historical continuity. Note that the target approximations $\vec{\beta}_k^{(j)}$ are not reused either anywhere in the aggregation so as to ensure that no error gets inadvertantly amplified by a feedback loop.

Note that the above choices to exclude the targets from (most) aggregations are not founded on an absolute rationale. They are merely precautions against overfitting and error amplification. One may very well devise regularization mechanisms that will alleviate these concerns and efficiently incorporate target histories as meaningful features in themselves.

A final observation is that not all three updates---i.e., of targets, intrinsic, and extrinsic features---systematically happen at every time step. Whenever an update is ``missing'', one shall not replace it with a null value, but simply carry over the last update together with its timestamp. Here again, this is not an absolute requirement as one could adopt alternative approaches, where missing values can be represented by dedicated values. Such design choices are mong those that require experimentation with a particular use case (cf. \S\ref{sec:Outlook}).

\section{Outlook}
\label{sec:Outlook}

In the broader context of ontologies, be them social, man-made, or natural, applications of machine learning have mostly been cross-sectional in that they deal with the estimation of a specific entity type only. This work attempts to further the reach of machine learning to arbitrary ontologies where the entities are attributed, dynamic, heterogeneous, and---more importantly---interacting with each other in ways that do not necessarily fit traditional graph topographies made up of bipartite edges. Estimation---i.e., classification or regression---is therefore not constrained to any given entity type anymore but can be applied accross all heterogeneous entities. Examples of such ontologies are illustrated in Fig. \ref{fig:ontologies}.

\begin{figure}
{\small
	\begin{subfigure}{1\textwidth}
		\begin{tabular}{ll}
			\multicolumn{2}{l}{\sc Cybersecurity}\vspace{7pt}\\
		 	& \begin{tabularx}{\textwidth}{lXXX}
		 	\toprule
		 	\textbf{entity} & \textbf{intrinsic} & \textbf{extrinsic} & \textbf{belief} \\
		 	\toprule
		 	e-mail & date sent; date received; text in the body; etc. & IP address of the sending server; DMARC policy; SPF policy; attached files and their attributes; reply-to e-mail address; etc. & benign; phishing; spam; malware; etc. \\\\
		 	IP address & autonomous system; geographical location; etc. & domain names hosted at that IP; registrants of those domains; active ports; etc. & benign; hosts a particular piece of malware; etc. \\\\
		 	file & size in bytes; hash signature; metadata; etc. & location in the hosting device; permissions; name of the author; etc. & benign; virus; spyware; ransomware; etc. \\
		 	etc $\ldots$ & & \\
		 	\bottomrule
		 	\end{tabularx}\\
		 	&  \\
		 	& \textbf{Interactions:} an e-mail is sent; a domain name is queried; a file is opened; a domain name is registered; etc.
		\end{tabular}
	\caption{Onotlogy of cybersecurity}
	\label{fig:cybersecurity ontology}
	\end{subfigure}\vspace{10pt}\\
	\begin{subfigure}{1\textwidth}
		\begin{tabular}{ll}
			\multicolumn{2}{l}{\sc Disease spread}\vspace{7pt}\\
		 	& \begin{tabularx}{\textwidth}{lXXX}
		 	\toprule
		 	\textbf{entity} & \textbf{intrinsic} & \textbf{extrinsic} & \textbf{belief} \\
		 	\toprule
		 	human & age; genetic predisposition; co-morbidity; etc. & walk of life; sociability; medical insurance; etc. & virus-free; asymptomatic carrier; at risk; etc. \\\\
		 	animal & genetic predisposition; etc. & proximity to humans; etc. & carrier; virus-free; etc. \\\\
		 	object & surface area; surface temperature; humidity; etc. & public; private; shared; frequency of disinfection; etc. & deposited with the virus; virus-free; etc. \\\\
		 	venue & capacity; ventilation; room temperature; etc. &  public; private; shared; frequency of disinfection;  & hot-spot for infection; virus-free; etc. \\
		 	etc $\ldots$ & & \\
		 	\bottomrule
		 	\end{tabularx}\\
		 	&  \\
		 	& \textbf{Interactions:} several people share the same object; an animal is sold at a market; two people shake hands; etc.
		\end{tabular}
	\caption{Ontology of disease spread}
	\label{fig:disease ontology}
	\end{subfigure}
}
\caption{Examples of ontologies and their corresponding building blocks for the purposes of supervised machine learning.}
\label{fig:ontologies}
\end{figure}

The reference architecture presented herein is intentially kept as high-level as possible, thereby allowing for the modular implementation of the four aggregators in a way that is agnostic as to their inner components, be them neural networks or any other technique. Given any particular problem domain, further investigation is therefore needed as to the particular design of the aggregators, in particular when it comes to such issues as
\begin{itemize}
\item the initialization of the latent representations $\hatf_k^{(j)}(t=0)$ and $\hatchi_k^{(j, i)}(t=0)$,
\item the choice of the mappings $\Map^{(j)}$, $\Map_f^{(j)}$, $\Map_{\chi}^{(j, i)}$, $\chi_l^{(i)}$ and their respective hyperparameters,
\item the initialization of the mappings via transfer learning, whenever applicable, 
\item the handling of missing values for $\vecf_k^{(j,i)}(t)$, $\vecb_k^{(j)}(t)$ or $\vecchi_k^{(j,i)}(t)$ at any given time step $t$,
\item the validity of the Markovian assumption in the time-aggregators, or
\item the choice of training scheme (batch vs. online).
\end{itemize}

\begin{ack}
This work was supported by the Innovation Fund Denmark. The author would like to thank Egon Kidmose for valuable feedback on the manuscruipt.
\end{ack}

\bibliographystyle{plainnat}
\bibliography{references}

\begin{thebibliography}{12}
\providecommand{\natexlab}[1]{#1}
\providecommand{\url}[1]{\texttt{#1}}
\expandafter\ifx\csname urlstyle\endcsname\relax
  \providecommand{\doi}[1]{doi: #1}\else
  \providecommand{\doi}{doi: \begingroup \urlstyle{rm}\Url}\fi

\bibitem[{Al-Eidi} et~al.(2020){Al-Eidi}, {Chen}, {Darwishand}, and
  {Alfosool}]{al-eidi2020time-ordered}
S.~{Al-Eidi}, Y.~{Chen}, O.~{Darwishand}, and A.~M.~S. {Alfosool}.
\newblock Time-ordered bipartite graph for spatio-temporal social network
  analysis.
\newblock In \emph{2020 International Conference on Computing, Networking and
  Communications (ICNC)}, pages 833--838, 2020.

\bibitem[Data61(2018)]{StellarGraph}
CSIRO's Data61.
\newblock Stellargraph machine learning library.
\newblock \url{https://github.com/stellargraph/stellargraph}, 2018.

\bibitem[Hamilton et~al.(2017)Hamilton, Ying, and
  Leskovec]{hamilton2017inductive}
William~L. Hamilton, Rex Ying, and Jure Leskovec.
\newblock Inductive representation learning on large graphs.
\newblock In \emph{Proceedings of the 31st International Conference on Neural
  Information Processing Systems}, NIPS’17, page 1025–1035, Red Hook, NY,
  USA, 2017. Curran Associates Inc.
\newblock ISBN 9781510860964.

\bibitem[Jiang et~al.(2019)Jiang, Wei, Feng, Cao, and Gao]{jiang2019dynamic}
Jianwen Jiang, Yuxuan Wei, Yifan Feng, Jingxuan Cao, and Yue Gao.
\newblock Dynamic hypergraph neural networks.
\newblock pages 2635--2641, 08 2019.
\newblock \doi{10.24963/ijcai.2019/366}.

\bibitem[Liu et~al.(2018)Liu, Chen, Yang, Zhou, Li, and
  Song]{liu2019heterogeneous}
Ziqi Liu, Chaochao Chen, Xinxing Yang, Jun Zhou, Xiaolong Li, and Le~Song.
\newblock Heterogeneous graph neural networks for malicious account detection.
\newblock In \emph{Proceedings of the 27th ACM International Conference on
  Information and Knowledge Management}, CIKM ’18, page 2077–2085, New
  York, NY, USA, 2018. Association for Computing Machinery.
\newblock ISBN 9781450360142.
\newblock \doi{10.1145/3269206.3272010}.
\newblock URL \url{https://doi.org/10.1145/3269206.3272010}.

\bibitem[Sun et~al.(2019)Sun, Tong, and Yang]{sun2019hindom}
Xiaoqing Sun, Mingkai Tong, and Jiahai Yang.
\newblock Hindom: A robust malicious domain detection system based on
  heterogeneous information network with transductive classification, 2019.

\bibitem[Vaswani et~al.(2017)Vaswani, Shazeer, Parmar, Uszkoreit, Jones, Gomez,
  Kaiser, and Polosukhin]{vaswani2017attention}
Ashish Vaswani, Noam Shazeer, Niki Parmar, Jakob Uszkoreit, Llion Jones,
  Aidan~N. Gomez, Lukasz Kaiser, and Illia Polosukhin.
\newblock Attention is all you need, 2017.

\bibitem[{Wu} et~al.(2020){Wu}, {Pan}, {Chen}, {Long}, {Zhang}, and
  {Yu}]{wu2020comprehensive}
Z.~{Wu}, S.~{Pan}, F.~{Chen}, G.~{Long}, C.~{Zhang}, and P.~S. {Yu}.
\newblock A comprehensive survey on graph neural networks.
\newblock \emph{IEEE Transactions on Neural Networks and Learning Systems},
  pages 1--21, 2020.
\newblock ISSN 2162-2388.
\newblock \doi{10.1109/TNNLS.2020.2978386}.

\bibitem[Yedidia et~al.(2003)Yedidia, Freeman, and
  Weiss]{yedida2003understanding}
Jonathan~S. Yedidia, William~T. Freeman, and Yair Weiss.
\newblock \emph{Understanding Belief Propagation and Its Generalizations}, page
  239–269.
\newblock Morgan Kaufmann Publishers Inc., San Francisco, CA, USA, 2003.
\newblock ISBN 1558608117.

\bibitem[Zhang et~al.(2018{\natexlab{a}})Zhang, Shi, Xie, Ma, King, and
  Yeung]{zhang2018gaan}
Jiani Zhang, Xingjian Shi, Junyuan Xie, Hao Ma, Irwin King, and Dit-Yan Yeung.
\newblock Gaan: Gated attention networks for learning on large and
  spatiotemporal graphs.
\newblock 03 2018{\natexlab{a}}.

\bibitem[Zhang et~al.(2018{\natexlab{b}})Zhang, Cui, and Zhu]{zhang2018deep}
Ziwei Zhang, Peng Cui, and Wenwu Zhu.
\newblock Deep learning on graphs: A survey, 2018{\natexlab{b}}.

\bibitem[Zhou et~al.(2006)Zhou, Huang, and Sch\"{o}lkopf]{zhou2006learning}
Dengyong Zhou, Jiayuan Huang, and Bernhard Sch\"{o}lkopf.
\newblock Learning with hypergraphs: Clustering, classification, and embedding.
\newblock In \emph{Proceedings of the 19th International Conference on Neural
  Information Processing Systems}, NIPS’06, page 1601–1608, Cambridge, MA,
  USA, 2006. MIT Press.

\end{thebibliography}
\end{document}